%% file: main.tex
\documentclass[journal]{IEEEtran}

\usepackage{times}
\usepackage{epsfig}
\usepackage{graphicx}
\usepackage{amsmath,amssymb,amsfonts}
\usepackage{algpseudocode}
\usepackage{booktabs}
\usepackage[pagebackref,breaklinks,colorlinks]{hyperref}
\newcommand{\TODO}[1]{\textbf{\color{red}[TODO: #1]}}
\renewcommand{\TODO}[1]{}

\usepackage[utf8]{inputenc}
\usepackage[T1]{fontenc}
\usepackage{multirow}
\usepackage{array}
\usepackage{makecell}
\usepackage{tabularx}
\usepackage{colortbl}
\usepackage{xcolor}
\usepackage{hhline} 
\usepackage[ruled,vlined]{algorithm2e}
\definecolor{myblue}{rgb}{0,0.2,0.6}

\newcommand{\thickhline}{\Xhline{2\arrayrulewidth}}
\definecolor{lightgray}{gray}{.95}

\usepackage{bm}
\usepackage{multirow}
\usepackage{wrapfig}
\usepackage[export]{adjustbox}

\usepackage{colortbl}
\definecolor{lightgray}{gray}{.9}
\definecolor{deepgray}{gray}{.8}

\usepackage{threeparttable}
\newcolumntype{I}{!{\vrule width 1pt}}
\definecolor{mygray}{gray}{.95}
\usepackage{float}
\usepackage[ruled, vlined]{algorithm2e}
\usepackage{pifont}
\usepackage{ragged2e}
\usepackage{tcolorbox}

\definecolor{DarkBlue}{RGB}{64,101,149} 

\definecolor{myblue}{RGB}{100, 115, 160}

\definecolor{DarkGreen}{RGB}{42,110,63}
\newtcbox{\tcbhighmath}[1][]{on line, colframe=white,
  colback=yellow!20, boxrule=0pt, arc=3pt, boxsep=0pt, left=3pt, right=3pt, top=3pt, bottom=3pt, #1}

\begin{document}

\title{FedThief: Harming Others to Benefit Oneself in Self-Centered Federated Learning}

\author{Xiangyu Zhang,
        Mang Ye,~\IEEEmembership{Senior Member, IEEE}
       % <-this % stops a space

\thanks{Xiangyu Zhang is with Key Laboratory of Aerospace Information Security and Trusted Computing, Ministry of Education, School of Cyber Science and Engineering, Wuhan University, Wuhan 430072, China (e-mail: xiangyuzhang@whu.edu.cn).}
\thanks{Mang Ye is with the School of Computer Science, Taikang Center for
Life and Medical Sciences, Wuhan University, Wuhan 430071, China (e-mail:
yemang@whu.edu.cn).}
\thanks{
% Manuscript received June 26, 2025; revised June XX, 2025.
Corresponding author: Mang Ye}% <-this % stops a space

% <-this % stops a space
% \thanks{Corresponding author: Mang Ye (yemang@whu.edu.cn).}
}

\maketitle

\input{./sec/0_abstract}


\IEEEpeerreviewmaketitle

\input{./sec/1_intro}
\input{./sec/2_related_work}
\input{./sec/3_method}
\input{./sec/4_experiment}
\input{./sec/5_conclusion}

% \appendices
% \section{Proof of Theorem 1}
% Appendix content here.

% \section*{Acknowledgment}

\bibliographystyle{IEEEtran}
\bibliography{main}

% you can use \bibliographystyle{IEEEtran} and \bibliography{} if needed

\end{document}

%% file: sec/0_abstract.tex
\begin{abstract}
In federated learning, participants’ uploaded model updates cannot be directly verified, leaving the system vulnerable to malicious attacks. Existing attack strategies have adversaries upload tampered model updates to degrade the global model’s performance. However, attackers also degrade their own private models, gaining no advantage.
In real-world scenarios, attackers are driven by self-centered motives: their goal is to gain a competitive advantage by developing a model that outperforms those of other participants, not merely to cause disruption.
In this paper, we study a novel Self-Centered Federated Learning (SCFL) attack paradigm, 
in which attackers not only degrade the performance of the global model through attacks but also enhance their own models within the federated learning process.
We propose a framework named FedThief, which degrades the performance of the global model by uploading modified content during the upload stage. At the same time, it enhances the private model's performance through divergence-aware ensemble techniques—where “divergence” quantifies the deviation between private and global models—that integrate global updates and local knowledge.
 Extensive experiments show that our method effectively degrades the global model performance while allowing the attacker to obtain an ensemble model that significantly outperforms the global model.
\end{abstract}

%% file: sec/1_intro.tex
\section{Introduction}
\label{sec:intro}

\IEEEPARstart{I}{n} the field of machine learning, the quality and diversity of the training data are widely recognized as essential prerequisites for enabling models to generalize effectively to unseen data and perform reliably across a range of downstream tasks \cite{zhou2017machine, zhou2024personalized}. These characteristics directly influence the learned model's empirical risk minimization, hypothesis space coverage, and robustness to distributional shifts \cite{rongcan, li2025boosting}. However, in many real-world scenarios, collecting such high-quality, heterogeneous data poses formidable challenges. Specifically, data acquisition is often constrained by numerous practical and ethical considerations, including cost constraints, institutional barriers, and data privacy concerns \cite{lameski2019challenges, bai2023text}.

This dilemma is particularly acute for individual-level organizations or isolated entities, who may lack the necessary resources or legal capacity to aggregate sufficient quantities of useful training data\cite{rashid2025trustworthy}. Furthermore, many domain-specific datasets—especially those involving personally identifiable information (PII), medical records, or financial transactions—contain highly sensitive or confidential data \cite{wu2022communication, dimitrov2016medical}. These characteristics impose stringent regulations on data sharing, governed by laws such as GDPR \cite{li2019impact}, HIPAA \cite{gostin2009beyond}, or regional data compliance mandates. As a result, naively centralizing data for joint model training becomes both legally and technically infeasible \cite{hukkeri2025split, zhao2025federation}.

To address these challenges, Federated Learning (FL) has emerged as a decentralized privacy-preserving machine learning paradigm that facilitates collaborative training among multiple distributed data owners without requiring raw data exchange \cite{mcmahan2017communication, zhang2021survey, huang2024federated}. FL allows all participants—commonly referred to as clients—to train a shared model collaboratively, while only sharing intermediate model updates such as gradients or parameters computed over local datasets \cite{fu2023client, fang2022robust}. This paradigm significantly alleviates privacy and security risks, and offers a promising alternative for data-isolated systems where collaborative modeling is desirable but direct information interchange is prohibited \cite{yazdinejad2024robust}.

\begin{figure*}[ht]
    \centering
    \setlength{\abovecaptionskip}{-10pt}
    \setlength{\belowcaptionskip}{-10pt}
    \includegraphics[width=1.0\textwidth]{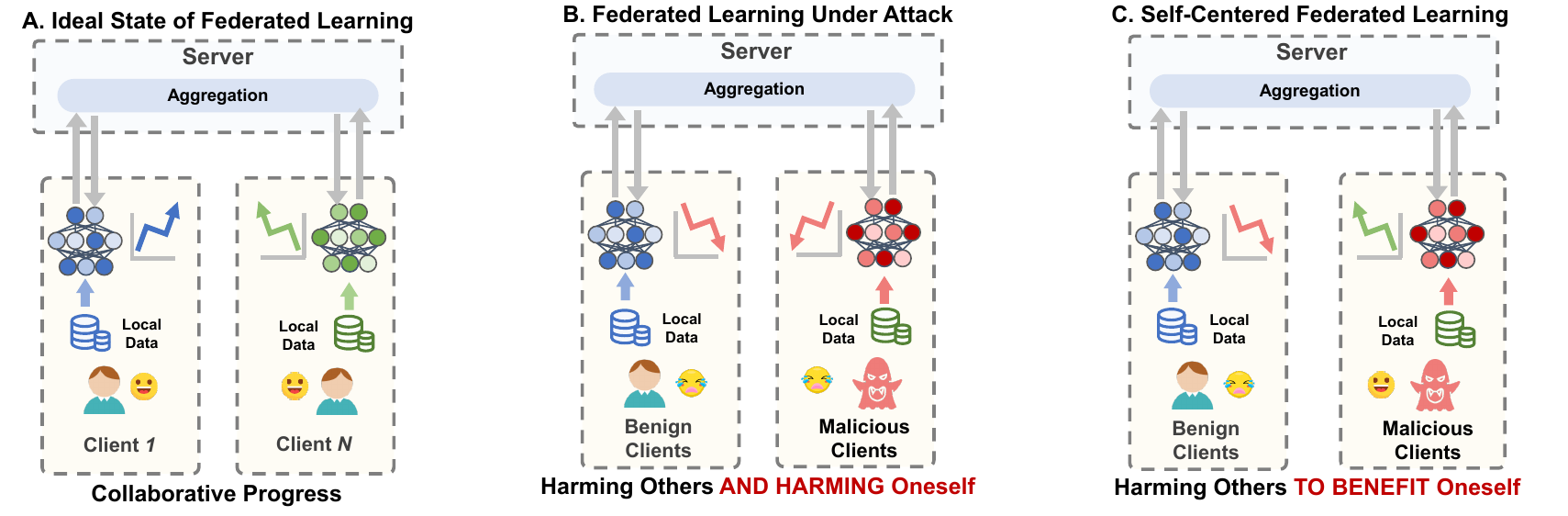}
    \caption{\textbf{Framework of Self-Centered Federated Learning. }
   \textbf{(A)}~Ideal State of Federated Learning: Multiple clients sharing corresponding feedback collectively train a high-performance model; 
   \textbf{(B)}~Federated Learning Under Attack: Malicious clients upload incorrect content to degrade the global model, but they do not benefit themselves; 
   \textbf{(C)}~Self-Centered Federated Learning: Malicious clients upload incorrect parameters and locally recalculate the correct parameters to obtain a high-performance private model.
}
    \label{fig:Three FL methods}
\end{figure*}

Within the federated learning (FL) framework, participating entities, referred to as clients, independently maintain their own datasets and locally train private models on this proprietary information~\cite{fang2024byzantine}. To ensure data locality and minimize privacy leakage, local model training is executed in isolation, and only derived statistics—such as parameter updates or gradients—are periodically shared with a central server \cite{pei2024review}. This server, in turn, employs a predefined aggregation strategy (e.g., Federated Averaging\cite{mcmahan2017communication}) to synthesize a global model that integrates knowledge from all participating clients. The updated global model is then broadcast back to the clients for further fine-tuning on their respective local datasets~\cite{qi2024model, ye2023heterogeneous}. This iterative optimization process is repeated over multiple communication rounds until the global model converges to satisfactory performance~\cite{zhang2022challenges}. Throughout this process, raw data never leaves the client’s device or administrative domain, thereby offering a strong inherent privacy guarantee~\cite{huang2022learn} and making FL fundamentally suitable for privacy-preserving machine learning, particularly in sensitive application scenarios, as illustrated in Fig.~\ref{fig:Three FL methods}(A).
Despite the privacy-preserving nature of federated learning (FL), the indirect exchange of information—via intermediate model updates rather than raw data—introduces significant security vulnerabilities\cite{bai2025provfl, xhemrishi2025fedgt}. Specifically, since the central server lacks direct access to client-side data, it cannot reliably verify whether the received updates are legitimate or adversarial~\cite{huang2024self, boenisch2023curious}. In many practical FL deployments—particularly those spanning business alliances, consortia, or industrial coalitions—clients may act as both collaborators and competitors~\cite{lyu2020threats}. As such, the existence of strategically malicious participants is not merely a theoretical concern, but a tangible and pervasive threat that can severely compromise collaborative model quality, consistency, and fairness.

% \begin{figure*}[ht]
%     \centering
%     \setlength{\abovecaptionskip}{-10pt}
%     \setlength{\belowcaptionskip}{-10pt}
%     \includegraphics[width=1.0\textwidth]{IEEE/Pic/Motivation_v4.pdf}
%     \caption{\textbf{Framework of Self-Centered Federated Learning. }
%    \textbf{(A)}~Ideal State of Federated Learning: Multiple clients sharing corresponding feedback collectively train a high-performance model; 
%    \textbf{(B)}~Federated Learning Under Attack: Malicious clients upload incorrect content to degrade the global model, but they do not benefit themselves; 
%    \textbf{(C)}~Self-Centered Federated Learning: Malicious clients upload incorrect parameters and locally recalculate the correct parameters to obtain a high-performance private model.
% }
%     \label{fig:Three FL methods}
% \end{figure*}

Existing literature confirms that malicious clients can conduct Byzantine attacks, a class of adversarial behavior in which compromised clients upload erroneous, misleading, or adversarial updates to the server during aggregation \cite{zhou2021deep, lyu2020threats, shejwalkar2021manipulating, fang2020local, tolpegin2020data}. These attacks degrade the performance of the global model, reduce convergence speed, or destabilize the training process. However, conventional Byzantine attack frameworks largely emphasize disrupting collaboration by reducing the overall utility of the system without necessarily enabling tangible benefits for the attackers themselves\cite{li2024efficiently, dai2024decaf}. In other words, malicious participants incur the cost of training and communication in order to sabotage the global model, but such sabotage often leads to degraded performance of their own private models as well—see Figure \ref{fig:Three FL methods}(B).

Therefore, conventional attack designs that solely focus on degrading the global model fail to reflect a crucial aspect of real-world adversarial behavior: the self-serving nature of malicious participants. In most practical threat models, attackers are utility-maximizing rational agents whose primary objective is not merely to harm others, but to establish a competitive advantage by securing superior model performance for themselves. For example, in financial or medical domains, gaining access to a better model than other participants can result in direct economic or strategic incentives\cite{lee2023federated, pourroostaei2023federated}. This observation leads us to a critical insight: a significant discrepancy exists between existing attack models and the self-interested motivations observed in practical settings. Thus, it is essential to explore alternative attack paradigms that explicitly couple harm to others with benefit to self.

In this paper, we propose a novel attack paradigm termed Self-Centered Federated Learning (SCFL). In SCFL, the attacker’s objective shifts from mere sabotage to gain: malicious clients covertly participate in the federated learning process and exploit it to develop private models that substantially outperform the globally aggregated model shared among all participants—Figure \ref{fig:Three FL methods}(C). This is accomplished through a two-phase strategy. In the first phase, during local training, attackers perform carefully designed Byzantine-style manipulations to upload distorted updates that lower the quality of the global model. In the second phase, after receiving the aggregated global updates, attackers leverage their prior knowledge of the previously-injected malicious components to strategically reverse their effects through correction or filtering, thereby selectively extracting useful information while sustaining performance gains for their private models.

However, implementing the SCFL paradigm is associated with several unique and practical challenges. Firstly, the server's aggregation mechanism is typically unknown or non-transparent to clients, making it difficult to explicitly quantify adversarial contributions or reliably anticipate the exact form of the aggregated global model. Moreover, in order to craft a superior private model, adversarial clients must intentionally diverge from the optimization trajectory of the global model during local training. As a result, the cumulative divergence between the private and global models tends to increase steadily over communication rounds, potentially amplifying model inconsistencies. This growing divergence raises two tightly coupled challenges.

On one hand, a widening divergence increases the detectability of manipulated updates. That is, as the attacker’s model deviates further in representation space from the global consensus, the server may flag these updates as anomalous or nonconforming, triggering potential countermeasures. On the other hand, increasing divergence reduces the utility of subsequent global updates distributed by the server, making them no longer directly applicable for enhancing the private model. As such, attackers face the dual challenge of remaining stealthy while optimizing for their own performance. Hence, designing a practical and robust mechanism that enables malicious agents to continually train superior private models while remaining undetected constitutes a core technical barrier.

To address these technical difficulties, we propose \textit{FedThief}—a general framework for constructing SCFL attacks. The design of FedThief is centered on a dual-model architecture, accommodating both attack execution and self-benefit extraction. Specifically, malicious clients simultaneously maintain and train:

A malicious model that stays aligned with the global model's training trajectory. This model serves to fabricate adversarial updates for submission to the server, ensuring that the attacker avoids excessive divergence and maintains stealth;
An ensemble-based private model \cite{dong2020survey, sagi2018ensemble, dey2023ensemble} that leverages divergence-aware optimization strategies. This model integrates three sources of knowledge: (i) the attacker’s own private update direction; (ii) feedback from global model updates; and (iii) synthesized corrections for previously-injected adversarial components. This ensemble boosts performance while maintaining a camouflage effect.
We summarize our main contributions as follows:

• We propose \textbf{Self-Centered Federated Learning (SCFL)}, a new and practically motivated federated attack paradigm that prioritizes both harming global model integrity and boosting local model performance for malicious clients.

• We introduce \textbf{FedThief}, a novel dual-model attack strategy that enables malicious clients to balance stealth and private gains via divergence-aware ensemble learning, deviating from the global model just enough to remain undetected while consistently maximizing private performance.

• We conduct comprehensive experimental evaluations on multiple benchmarks, which demonstrate that malicious clients using FedThief can consistently obtain private models that outperform the global model in both accuracy and convergence speed, thus empirically validating the practical threat of SCFL in realistic federated learning deployments.

% The rest of the paper is structured as follows. In Section~\ref{sec:Related Work}, we review related work on federated learning security, with emphasis on model poisoning attacks and defense mechanisms. In Section~\ref{sec:Method}, We present the proposed method, \textit{FedThief}, describing its overall framework, self-centered attack execution strategy, model-specific local update mechanism, and divergence-aware ensemble optimization procedure. Then, we report experimental results in Section~\ref{sec:Experiments} to validate the effectiveness and robustness of FedThief under various attack scenarios. Finally, Section~\ref{sec:Conclusion} concludes the paper and outlines potential directions for future research.

%% file: sec/2_related_work.tex
\section{Related Work}
\label{sec:Related Work}
\subsection{Byzantine Attacks in Federated Learning}
Byzantine attacks in federated learning (FL) aim to subvert the global model either by corrupting local data or directly manipulating model updates submitted to the server. These attacks are generally categorized into two types: data-poisoning attacks and model- (or gradient-) poisoning attacks.

\paragraph{Data-Poisoning Attacks}  
In data-poisoning attacks, adversarial clients poison their local training data with the intent of degrading global model performance on benign inputs. Early work by Van~\emph{et al.}~\cite{van2015learning} introduces \emph{Symmetry Flipping}, in which class labels are randomly reassigned with equal probability, thereby injecting label noise that disrupts convergence. Han~\emph{et al.}~\cite{han2018co} propose \emph{Pair Flipping}, where labels are replaced with semantically similar classes to induce targeted misclassification. More recently, Liu~\emph{et al.}~\cite{liu2024badsampler} develop \emph{BadSampler}, a clean-label attack method in which malicious clients selectively sample high-loss data points during local training, steering the model toward higher generalization error without altering any ground-truth labels.

A notable subclass of data-poisoning is \textit{backdoor attacks}, which preserve benign performance while embedding malicious behavior triggered by specific inputs. Sun~\emph{et al.}~\cite{sun2019can} and Zhang~\emph{et al.}~\cite{zhang2023backdoor} provide surveys of backdoor injection techniques in FL. Bagdasaryan~\emph{et al.}~\cite{bagdasaryan2020backdoor} demonstrate that such attacks can succeed even under robust aggregation. Xie~\emph{et al.}~\cite{xie2019dba} propose \emph{DBA}, which dynamically adjusts trigger strength across clients to enhance stealth and effectiveness.

\paragraph{Model-Poisoning Attacks}  
In model-poisoning attacks, adversaries craft malicious updates to manipulate the aggregation result at the server. Baruch~\emph{et al.}~\cite{baruch2019little} introduce the \emph{LIE} attack, where the gradient $g_m$ from each malicious client is perturbed by random noise drawn from a Gaussian distribution proportional to the per-coordinate standard deviation $\sigma$ of the benign gradients:
\begin{equation}
  \tilde g_m = g_m + z,\quad z_i \sim \mathcal{N}(0, \alpha\,\sigma_i).
\end{equation}
This low-variance perturbation allows malicious contributions to bypass defenses such as Trimmed-Mean and Median. Shejwalkar and Houmansadr~\cite{shejwalkar2021manipulating} propose \emph{Min-Sum}, where the manipulated gradient satisfies
\begin{equation}
  \sum_{j=1}^N \| \tilde g_m - g_j \|^2 \le \tau,
\end{equation}
ensuring that the adversarial update remains within a Euclidean ball centered around the benign client cluster. More recently, Ma~\emph{et al.}~\cite{ma2025fedghost} introduce \emph{FedGhost}, which dynamically estimates the adversarial contribution in server-side aggregation and adaptively adjusts the attack magnitude in real-time, thereby maximizing disruption while effectively evading detection by standard statistical filters.

\subsection{Defense Strategies Against Byzantine Attacks}
To mitigate the impact of Byzantine attacks, a variety of defense strategies have been developed. These include distance-based filtering, statistically robust aggregation, and the use of proxy datasets at the server. We categorize existing methods into three groups: distance-based defenses, statistical aggregation defenses, and proxy-dataset-based defenses.

\subsubsection{Distance-Based Defenses}
These methods identify and suppress potentially malicious updates by quantifying their deviation from the majority consensus.

\paragraph{Krum and Multi-Krum}  
Blanchard~\emph{et al.}~\cite{blanchard2017machine} propose \emph{Krum}, which selects the update with the minimal sum of squared distances to its $N - f - 2$ nearest neighbors:
\begin{equation}
  \mathrm{score}(g_i) = \sum_{j \in \mathcal{N}_i} \| g_i - g_j \|_2^2,
\end{equation}
where $\mathcal{N}_i$ denotes the set of nearest gradients. \emph{Multi-Krum} generalizes this by selecting the top-$m$ lowest-score updates and averaging them:
\begin{equation}
  g_{\mathrm{multi}} = \frac{1}{m} \sum_{i \in \mathcal{S}_m} g_i.
\end{equation}

\paragraph{FoolsGold}  
Fung~\emph{et al.}~\cite{fung2018mitigating} address coordinated collusion among malicious clients by computing pairwise cosine similarity between updates,
\begin{equation}
  \cos(g_i, g_j) = \frac{\langle g_i, g_j \rangle}{\| g_i \|_2\, \| g_j \|_2},
\end{equation}
and down-weighting clients with highly similar gradients:
\begin{equation}
  w_i \propto 1 - \max_{j \neq i} \cos(g_i, g_j), \quad
  g_{\mathrm{agg}} = \sum_{i=1}^N w_i g_i.
\end{equation}

\subsubsection{Statistical Aggregation Defenses}
These defenses use coordinate-wise robust statistics to suppress outliers.

\paragraph{Trimmed-Mean}  
Yin~\emph{et al.}~\cite{pmlr-v80-yin18a} propose \emph{Trimmed-Mean}, which discards the top $f$ and bottom $f$ values in each dimension and averages the remaining:
\begin{equation}
  [g_{\mathrm{trim}}]_d = \frac{1}{N - 2f} \sum_{i = f + 1}^{N - f} g_{(i),d},
\end{equation}
where $g_{(i),d}$ denotes the $i$-th order statistic for dimension~$d$.

\paragraph{Median}  
A simpler yet effective method applies the coordinate-wise median aggregation rule:
\begin{equation}
  [g_{\mathrm{med}}]_d = \mathrm{median}\{g_{1,d},\ldots,g_{N,d}\}.
\end{equation}

\paragraph{Bulyan}  
Guerraoui~\emph{et al.}~\cite{guerraoui2018hidden} introduce \emph{Bulyan}, which combines Multi-Krum and robust statistics. It first selects $2f+2$ gradients via Multi-Krum, and then applies Trimmed-Mean or Median aggregation rule within this subset to compute the final aggregated update $g_{\mathrm{bulyan}}$.

\subsubsection{Proxy-Dataset Defenses}
This class of methods assumes access to a small, trusted dataset residing at the server.

\paragraph{FLTrust}  
Cao~\emph{et al.}~\cite{cao2021fltrust} propose \emph{FLTrust}, where the server maintains a clean proxy dataset $D_{\mathrm{proxy}}$ to compute a reference gradient:
\begin{equation}
  g_{\mathrm{ref}} = \nabla \ell(\theta;\, D_{\mathrm{proxy}}),
\end{equation}
and uses cosine similarity to reweight each client's update:
\begin{equation}
  \alpha_i = \max(0,\,\cos(g_i, g_{\mathrm{ref}})), \quad
  g_{\mathrm{agg}} = \frac{ \sum_i \alpha_i g_i }{ \sum_i \alpha_i }.
\end{equation}
Clients with updates deviating from the server's reference receive lower weights.

%% file: sec/3_method.tex
\section{Proposed Method}
\label{sec:Method}
In this section, we first present the overall framework and then describe each core component in detail. The notations used throughout this section are summarized in Table~\ref{tab:notation}.

\begin{table}[t]
\caption{Summary of Notation}
\label{tab:notation}
\centering
\renewcommand{\arraystretch}{1.2}
\begin{tabular}{l||p{6.5cm}}
\hline\hline
\textbf{Symbol} & \textbf{Description} \\
\hline
$N$ & Total number of clients \\
$\mathcal{K},\, \mathcal{K}_m$ & All clients; subset of malicious clients \\
$C$ & Number of classes \\
$D_k^{\text{train}},\, D_k^{\text{val}}$ & Training and validation splits of client $c_k$ \\
$x_k^i,\, y_k^i$ & $i$-th input and one-hot label of client $c_k$ \\
$\theta_p,\, \theta_m,\, \theta_e$ & Parameters of private, malicious, and error models \\
$g_p^t,\, g_m^t$ & Private and malicious gradients at round $t$ \\
$\tilde{g}_m^t$ & Perturbed gradient after Byzantine attack \\
$\mathcal{A}(\cdot)$ & Byzantine attack function \\
$\delta,\, \beta$ & Perturbation vector and its magnitude \\
$g^t$ & Global aggregated gradient at round $t$ \\
$\eta,\, \lambda$ & Learning rate; loss trade-off weight \\
$\mathcal{L}_{\mathrm{CE}},\, \mathcal{L}_{\mathrm{KD}}$ & Cross-entropy and KL divergence losses \\
$\mathcal{E}_k^t,\, \mathcal{L}^t$ & Local ensemble head; global ensemble model \\
\hline\hline
\end{tabular}
\end{table}

\subsection{Overview of FedThief}

We consider a federated learning system designed for a \( C \)-class image classification task, consisting of  \( N \) clients and a central server.  Among these clients, an \( \alpha \) fraction of the clients is controlled by adversaries, referred to as malicious clients.
Each client \( c_k \) maintains a local private dataset \( D_k = \{(x_k^i, y_k^i)\}_{i=1}^{N_k} \), where \( x_k^i \in \mathbb{R}^{d_{\text{in}}} \) represents an input sample with \( d_{\text{in}} \) feature dimensions, and \( y_k^i \in \{0,1\}^C \) denotes the corresponding one-hot encoded label vector. The dataset size is given by \( |D_k| = N_k \).  

Each client trains a local model \( f(\theta_k) \), parameterized by \( \theta_k \in \mathbb{R}^{d_{\theta}} \), which maps input samples to class predictions:  

\begin{equation}
    f(\theta_k) : \mathbb{R}^{d_{\text{in}}} \to \mathbb{R}^C.
\end{equation}

Federated learning proceeds in an iterative manner, where clients upload local updates and the server aggregates these updates and redistributes the resulting global updates to synchronize the local models of the clients.
Benign clients contribute model updates based on local training, while malicious clients deliberately craft manipulated updates to disrupt the global learning process.  

\begin{figure*}[t!]
    \centering
    \includegraphics[width=0.95\textwidth]{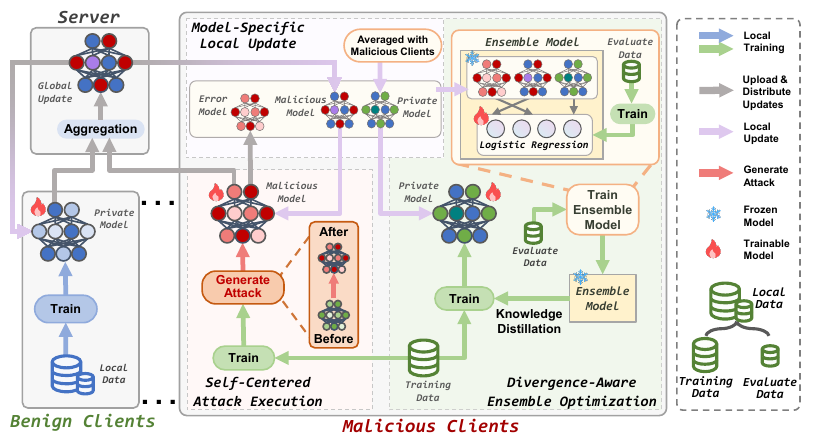}
    \caption{\textbf{Framework of the proposed FedThief approach.}Each malicious client maintains a private model for local training and a malicious model for adversarial attacks. 
     The private dataset is divided into training and validation subsets.
    During training, both models learn from the training set, where the malicious model generates manipulated gradients that are uploaded to the server.
    After receiving the global updates from the server, an ensemble model is constructed by integrating the private model, malicious model, and global updates. 
    A logistic regression model trained by the validation set is employed to dynamically combine the logits from these individual models, thereby generating the final predictions for the ensemble model.
    In the new training round, the ensemble model guides the private model's local training through knowledge distillation.} 
    \label{fig:framework}
\end{figure*}

% \subsection{Key Components of FedThief}

As shown in Figure \ref{fig:framework}, in FedThief, each malicious client independently maintains four local models:  

- \textbf{Private model} \( \theta_p \): optimized for local accuracy.

- \textbf{Malicious model} \( \theta_m \): used to craft adversarial updates.

- \textbf{Error model} \( \theta_e \): used to evaluate attack degradation.

- \textbf{Ensemble model} \( \mathcal{L} \): the final output model of the malicious client, which integrates the outputs of the private, malicious, and error models through a linear regression model\( \mathcal{E}_k^t \).

To facilitate the evaluation of the ensemble model's performance, a certain proportion \( \frac{1}{v} \) of the local dataset \( D_k \) is extracted as the validation set \( D_k^{\text{val}} \) used for training the linear regression model. while the remaining \( \frac{v-1}{v} \) are used as the training set \( D_k^{\text{train}} \).

\subsection{Self-Centered Attack Execution}

As shown in\ref{alg:attack}, for each malicious client at round \( t \), the malicious model \( \theta_m^t \) is trained independently on \( D_k^{\text{train}} \), and used solely for generating poisoned gradients.

The gradient of the malicious model is computed as:
\begin{equation}
\label{equ:1}
    g_m^t = \nabla f(\theta_m^t; D_k^{\text{train}}).
\end{equation}

\begin{algorithm}[ht]
\caption{Self-Centered Attack Execution}
\label{alg:attack}
\SetAlgoLined
\SetKwInput{KwInput}{Input}
\SetKwInput{KwOutput}{Output}
\SetNlSty{textbf}{}{}           
\DontPrintSemicolon   

\KwInput{
    Training data $D_k^{\text{train}}$; current malicious model parameters $\theta_m^t$; \\
    attack function $\mathcal{A}(\cdot)$; perturbation scale $\beta$
}
\KwOutput{Adversarial gradient $\tilde{g}_m^t$ uploaded to the server}

\textbf{1. Compute Local Gradient:} \\
$g_m^t \gets \nabla f(\theta_m^t; D_k^{\text{train}})$  \hfill \textit{\small // Eq.~\eqref{equ:1}} \;

\textbf{2. Apply Adversarial Perturbation:} \\
$\tilde{g}_m^t \gets \mathcal{A}(g_m^t) = g_m^t + \beta \cdot \delta$, where $\delta \in \mathbb{R}^{d_\theta}$ \hfill \textit{\small // Eq.~\eqref{equ:2}} \;

\textbf{3. Upload Poisoned Gradient:} \\
Send $\tilde{g}_m^t$ to the server \;

\vspace{1mm}
\textbf{4. [Server] Aggregate Global Update:} \\
$g^t \gets \text{Aggregate}(\{\tilde{g}_m^t\} \cup \{g_{\text{benign}}^t\})$  \hfill \textit{\small // Eq.~\eqref{equ:3}} \;

\end{algorithm}

A perturbation is applied to the malicious gradient to generate the adversarial gradient:
\begin{equation}
\label{equ:2}
    \tilde{g}_m^t = \mathcal{A}(g_m^t) = g_m^t + \beta \cdot \delta,
\end{equation}
where \( \mathcal{A}(\cdot) \) denotes the attack function, \( \delta \) is the adversarial direction, and \( \beta \in \mathbb{R}^+ \) is its magnitude.

The perturbed gradient \( \tilde{g}_m^t \) is uploaded to the server. Upon collecting updates from all clients, the server computes the global gradient:
\begin{equation}
\label{equ:3}
    g^t = \text{Aggregate}(\{\tilde{g}_m^t\} \cup \{g_{\text{benign}}^t\}).
\end{equation}

\subsection{Model-Specific Local Update}

After receiving the global gradient \( g^t \), each malicious client subsequently updates its local models accordingly.

\paragraph{Malicious model update}
The malicious model is synchronized with the global direction:
\begin{equation}
\label{equ:4}
    \theta_m^{t+1} = \theta_m^t - \eta g^t.
\end{equation}

\paragraph{Private model update}
Each malicious client computes the gradient for its private model using its local data:
\begin{equation}
\label{equ:5}
    g_p^t = \nabla f(\theta_p^t; D_k^{\text{train}}).
\end{equation}

The malicious clients in \( \mathcal{K}_m \) share their private gradients and perform collaborative local update:
\begin{equation}
\label{equ:6}
    \theta_p^{t+1} = \theta_p^t - \eta \cdot \frac{1}{|\mathcal{K}_m|} \sum_{k \in \mathcal{K}_m} g_{p,k}^t.
\end{equation}

\paragraph{Error model update}
The error model explicitly simulates global model degradation by being updated with the poisoned gradient.
\begin{equation}
\label{equ:7}
    \theta_e^{t+1} = \theta_e^t - \eta \cdot \tilde{g}_m^t.
\end{equation}
\subsection{Divergence-Aware Ensemble Optimization}

To effectively exploit the complementary characteristics of diverse internal model predictions, each malicious client constructs a lightweight ensemble classifier designed to aggregate multiple sources of locally derived knowledge. This divergence-aware strategy strengthens the client’s ability to generalize and adapt to malicious objectives without directly compromising individual model integrity, as shown in\ref{alg:ensemble}. 

\begin{algorithm}[t]
\caption{Divergence-Aware Ensemble Optimization for Malicious Client $c_k$}
\label{alg:ensemble}
\SetAlgoLined
\SetKwInput{KwInput}{Input}
\SetKwInput{KwOutput}{Output}
\SetNlSty{textbf}{}{}           
\DontPrintSemicolon   

\KwInput{
    Validation set $D_k^{\text{val}}$ with class labels $y$; 
    Local model parameters: $\theta_p^t$, $\theta_m^t$, $\theta_e^t$;
    Learning rate $\eta$; trade-off factor $\lambda$
}
\KwOutput{Updated private model parameters $\theta_p^{t+1}$}

\textbf{1. Logit Extraction on Validation Samples:} \;
\ForEach{$\mathbf{x} \in D_k^{\text{val}}$}{
    Compute logits:\;
    $z_p \gets f(\theta_p^t)(\mathbf{x})$\; 
    $z_m \gets f(\theta_m^t)(\mathbf{x})$\;
    $z_e \gets f(\theta_e^t)(\mathbf{x})$\;
}

\textbf{2. Train Local Ensemble Head $\mathcal{E}_k^t$:} \;
Learn weight composition via logistic regression: \\
$\min_{\mathcal{E}} \sum_{\mathbf{x} \in D_k^{\text{val}}} \mathcal{L}(z_m, z_p, z_e)$  \hfill \textit{\small // Eq.~\eqref{equ:8}} \;

\textbf{3. Global Ensemble Fusion:} \;
Aggregate all malicious clients' ensemble heads: \\
$\mathcal{L}^{t+1} \gets \frac{1}{|\mathcal{K}_m|} \sum_{k \in \mathcal{K}_m} \mathcal{E}_k^t$ \hfill \textit{\small // Eq.~\eqref{equ:9}} \;

\textbf{4. Compute Supervised Loss (Cross-Entropy):} \;
$\mathcal{L}_{\mathrm{CE}} \gets - \sum_{i=1}^{C} y^i \log f(\theta_p^{t+1})$ \hfill \textit{\small // Eq.~\eqref{equ:10}} \;

\textbf{5. Compute Knowledge Distillation Loss:} \;
$P \gets \sigma(z_{\text{ensemble}}), ~~ Q \gets \sigma(z_p)$ \;
$\mathcal{L}_{\mathrm{KD}} \gets \sum_i P_i \log \frac{P_i}{Q_i}$ \hfill \textit{\small // Eq.~\eqref{equ:11}} \;

\textbf{6. Combine Loss Components:} \;
$\mathcal{L}_{\mathrm{total}} \gets \lambda \cdot \mathcal{L}_{\mathrm{CE}} + (1 - \lambda) \cdot \mathcal{L}_{\mathrm{KD}}$ \hfill \textit{\small // Eq.~\eqref{equ:12}} \;

\textbf{7. Update Private Model via SGD:} \;
$\theta_p^{t+1} \gets \theta_p^{t+1} - \eta \cdot \nabla \mathcal{L}_{\mathrm{total}}$ \hfill \textit{\small // Eq.~\eqref{equ:13}} \;

\end{algorithm}

Let \( z_m \), \( z_p \), and \( z_e \) denote the predicted logits corresponding to the malicious model \( f(\theta_m^t) \), the private task model \( f(\theta_p^t) \), and an optional error modeling head \( f(\theta_e^t) \), respectively, all evaluated over the client’s local validation dataset \( \mathbf{x} \in D_k^{\text{val}} \). Based on these predictions, a local ensemble classifier \( \mathcal{E}_k^t \) is trained to combine the model outputs into a unified decision-making function. Specifically, this ensemble is realized via multinomial logistic regression, which facilitates a learnable convex combination over the logit space. 

The ensemble model is trained via multinomial logistic regression to minimize the cross-entropy loss between predictions and ground-truth labels, with L2 regularization to mitigate overfitting. The training is performed using L-BFGS until convergence.  The optimization problem is formally defined as:

\begin{equation}
\label{equ:8}
    \min_{\mathcal{E}} \sum_{\mathbf{x} \in D_k^{\text{val}}} \mathcal{L}(z_m, z_p, z_e),
\end{equation}

where \( \mathcal{L}(\cdot) \) denotes the regularized multi-class cross-entropy loss. This framework allows each malicious participant to extract synergistic utility from distinct prediction modes—capturing both adversarial trends and task-aligned representations—even in the presence of data heterogeneity or compromised model updates.

Upon completion of local ensemble optimization, each malicious client shares their logistic head \( \mathcal{E}_k^t \) with the rest of the malicious cohort. A centralized fusion process is then employed to aggregate these models into a unified global ensemble classifier \( \mathcal{L}^{t+1} \), which acts as a consensus soft teacher. This fusion is realized through a simple yet effective model averaging scheme:

\begin{equation}
\label{equ:9}
    \mathcal{L}^{t+1} = \frac{1}{|\mathcal{K}_m|} \sum_{k \in \mathcal{K}_m} \mathcal{E}_k^t,
\end{equation}

where \( \mathcal{K}_m \) denotes the set of all malicious clients. The resulting classifier \( \mathcal{L}^{t+1} \) is utilized as a soft-label generator for subsequent knowledge transfer into the client's private model, forming a crucial part of the targeted adaptation pipeline.
 
To preserve task-specific performance on authorized downstream tasks, each malicious client utilizes locally labeled data to provide a strong supervised learning signal. The traditional cross-entropy loss is computed between the model's predictions and the true class labels \( y \in \{1, \dots, C\} \), where \( C \) denotes the number of classes:

\begin{equation}
\label{equ:10}
    \mathcal{L}_{\mathrm{CE}} = - \sum_{i=1}^{C} y^i \log f(\theta_p^{t+1}),
\end{equation}

where \( f(\theta_p^{t+1}) \) denotes the softmax output of the updated private model on input \( \mathbf{x} \). This loss component ensures that the model remains well aligned with its primary utility task and effectively prevents performance deviation due to the integration of adversarial signals.

To incorporate the additional ensemble-based knowledge into the private model, we leverage knowledge distillation. Specifically, we compute the Kullback-Leibler (KL) divergence between the soft predictions from the global ensemble head \( z_{\text{ensemble}} \), and those generated by the private model \( z_p \). The distillation loss is expressed as:

\begin{equation}
\label{equ:11}
    \mathcal{L}_{\mathrm{KD}} = \sum_{i} P_i \log \frac{P_i}{Q_i},
\end{equation}

where \( P = \sigma(z_{\text{ensemble}}) \) and \( Q = \sigma(z_p) \) represent the softmax-normalized logit outputs of the ensemble head and private model, respectively. This mechanism facilitates the alignment of the private model's predictive distribution with the smoothed and consensus-informed distribution offered by the ensemble teacher.

To harness the benefits of both direct supervision and soft-label guidance, the training of the private model is driven by a composite loss function that balances the supervised learning signal and distilled ensemble knowledge. The total objective is formulated as:

\begin{equation}
\label{equ:12}
    \mathcal{L}_{\mathrm{total}} = \lambda \cdot \mathcal{L}_{\mathrm{CE}} + (1 - \lambda) \cdot \mathcal{L}_{\mathrm{KD}},
\end{equation}

where \( \lambda \in [0,1] \) is a balancing coefficient that controls the trade-off between standard supervised learning and cross-distribution alignment. The optimization step for updating the private model parameters \( \theta_p \) is then performed using standard stochastic gradient descent:

\begin{equation}
\label{equ:13}
    \theta_p^{t+1} = \theta_p^{t+1} - \eta \cdot \nabla \mathcal{L}_{\mathrm{total}},
\end{equation}

where \( \eta \) denotes the learning rate. This comprehensive update strategy enables the private model to steadily evolve across communication rounds, integrating both adversarially-informed ensemble knowledge and task-consistent labeled data to concurrently pursue high downstream performance and malicious objectives.

In summary, the proposed divergence-aware ensemble optimization serves as a critical component for harmonizing local learning dynamics and external knowledge transfer, thereby allowing malicious clients to discreetly exploit global model updates while steadily improving private task utility.

%% file: sec/4_experiment.tex
\section{Experimental Analysis}
\label{sec:Experiments}
% This section presents a comprehensive empirical analysis of the proposed FedThief framework. We divide our evaluations into two main parts. First, we conduct extensive \textit{comparison experiments}, focusing on the overall model performance in different federated learning configurations, involving varying datasets, attack strategies, aggregation defenses, and malicious client ratios. Second, we carry out an \textit{ablation study} to isolate and understand the effect of key parameters and components within FedThief, such as the loss balance factor, validation split, ensemble structure, and distillation hyperparameters.

\subsection{Experimental Setup}
\subsubsection{Datasets and Model Architectures}
To comprehensively evaluate the effectiveness of the proposed FedThief framework and ensure the validity and comparability of our results, we conduct experiments under standard image classification benchmarks and experimental configurations widely adopted in prior federated learning studies~\cite{huang2024federated,shejwalkar2021manipulating,ma2025fedghost}.
We select three widely used datasets of varying complexity: MNIST~\cite{lecun1998mnist}, CIFAR-10~\cite{krizhevsky2009learning}, and Fashion-MNIST (denoted as FASHION hereafter)~\cite{xiao2017fashion}.

Both MNIST and FASHION consist of ten-class grayscale images with resolution $28 \times 28$, whereas CIFAR-10 contains colored images of size $32 \times 32 \times 3$. All datasets are evenly and independently partitioned—i.e., under IID settings—across $50$ clients, ensuring class-balanced distributions without data heterogeneity. This allows us to focus exclusively on malicious behaviors and their direct impact, without the confounding factors introduced by non-IID local data.

For MNIST and FASHION, we implement a lightweight convolutional neural network (CNN) as the client model. It consists of three convolutional layers (kernel size: $3 \times 3$), each followed by ReLU activation, a $2 \times 2$ max-pooling layer, and three fully connected layers. For CIFAR-10, the client model is based on a variant of AlexNet~\cite{krizhevsky2009learning}, which integrates five convolutional layers, batch normalization, ReLU activations, and max-pooling, followed by dense layers.

All models are locally initialized and trained on clients. Updates are aggregated using standard federated optimization protocols described in Section~\ref{sec:Method}.

\subsubsection{Attack Method Settings}

In practical federated learning systems, malicious participants typically lack access to the precise aggregation algorithms or gradients shared by benign clients. To simulate such realistic threat models, we evaluate FedThief under three prominent Byzantine attack strategies that require neither knowledge of benign updates nor server aggregation rules:

\begin{itemize}
    \item \textbf{LIE}~\cite{baruch2019little} (Little is Enough): Injects a small but directionally consistent perturbation to scale malicious gradients.
    \item \textbf{MinSum}~\cite{shejwalkar2021manipulating}: Crafts updates to minimize the sum of distances to benign gradients.
    \item \textbf{FedGhost}~\cite{ma2025fedghost}: A query-free, decision-level attack using synthetic outputs instead of raw gradients.
\end{itemize}

To characterize the relationship between malicious participation and model resilience, we evaluate each method under two adversarial client ratios: $\alpha = 20\%$ and $\alpha = 40\%$.

\subsubsection{Defense Method Settings}

To benchmark the effectiveness of FedThief against existing learning paradigms and defenses, we compare our method with a variety of widely adopted FL baselines:

\begin{itemize}
    \item \textbf{FedAvg}~\cite{mcmahan2017communication}: The original and standard federated averaging algorithm used as a baseline.
    \item \textbf{FedProx}~\cite{li2020federated}: A variant of FedAvg that introduces a proximal term to handle client heterogeneity.
    \item \textbf{Median}, \textbf{Trimmed-Mean}~\cite{pmlr-v80-yin18a}: Robust aggregation rules based on coordinate-wise statistics.
    \item \textbf{Multi-Krum}~\cite{blanchard2017machine} and \textbf{Bulyan}~\cite{guerraoui2018hidden}: Byzantine-resilient methods that leverage geometric filtering to remove suspicious updates before aggregation.
\end{itemize}

\subsubsection{Evaluation Metrics}

Following the SCFL design, we independently evaluate both system-wide model performance and malicious gains. Specifically, we report:

\begin{itemize}
    \item \textbf{Global Model Accuracy ($Acc_g$)}: Classification accuracy of the aggregated global model in the final communication round, evaluated on a central held-out test set.
    \item \textbf{Malicious Ensemble Accuracy ($Acc_e$)}: Classification accuracy of the ensemble model locally constructed and used by malicious clients.
\end{itemize}

To quantify the stealthy efficacy of FedThief, we define a malicious advantage metric:
\begin{equation}
    \Delta_{\text{mal}} = Acc_e - Acc_g,
\end{equation}
which captures the extent to which malicious clients outperform the shared global model. For reference, we also report $\widetilde{Acc}_g$, the global model's nominal baseline accuracy under clean, attack-free settings.

\subsubsection{Implementation Details}

We configure the system with $50$ clients unless otherwise noted. Each client holds an equal partition of data, strictly under IID conditions. Federated training is performed over $40$ communication rounds, with default hyperparameters set as follows: learning rate $\eta = 0.001$, batch size = 256, and the Adam optimizer.

For local training, each client performs $2$ local epochs per global round on MNIST and FASHION, and $4$ epochs on CIFAR-10. For knowledge distillation in FedThief, we set the temperature hyperparameter $\tau = 3.0$, and the loss combination coefficient $\lambda = 0.5$.
All experiments are implemented using \texttt{PyTorch}, and conducted on NVIDIA RTX-3090 GPUs.

 \begin{table*}[t]\small
% \captionsetup{font=small}
\caption{The accuracy rates of the global model for benign clients (\(Acc_g\)) and the ensemble model for malicious clients (\(Acc_e\))  under a range of adversarial attacks and defensive strategies. \(\alpha\) represents the proportion of malicious clients.}
\label{tab:acc}
\centering
\scriptsize 
\resizebox{\linewidth}{!}{ 
\setlength\tabcolsep{2.pt}
\renewcommand\arraystretch{1.2}
\begin{tabular}{c|c|c||cccc|cccc|cccc}
\Xhline{1.2pt}

 \rowcolor[HTML]{f2f2f2}
&&&   \multicolumn{12}{c}{\textbf{Attack Method}}
\\
\cline{4-15}
 \rowcolor[HTML]{f2f2f2}
&&
&\multicolumn{4}{c|}{\cellcolor{white}\textbf{LIE} \cite{baruch2019little}}
&\multicolumn{4}{c|}{\cellcolor{white}\textbf{Min-Sum} \cite{shejwalkar2021manipulating}}
& \multicolumn{4}{c}{\cellcolor{white}\textbf{FedGhost} \cite{ma2025fedghost}}\\
\rowcolor[HTML]{f2f2f2}&&&\multicolumn{2}{c|}{$\alpha=0.2$}
& \multicolumn{2}{c|}{$ \alpha=0.4 $}
& \multicolumn{2}{c|}{$ \alpha=0.2 $}
& \multicolumn{2}{c|}{$ \alpha=0.4 $}
& \multicolumn{2}{c|}{$ \alpha=0.2 $}
& \multicolumn{2}{c}{$ \alpha=0.4 $}\\
\rowcolor[HTML]{f2f2f2}
\multirow{-4}{*}{\makecell{\textbf{Dataset}\\ \textbf{(Model)}}} 
& \multirow{-4}{*}{\makecell{\textbf{Aggregate}\\ \textbf{Methods}}} 
& \multirow{-4}{*}{\makecell{\textbf{No Attack}\\ \(\widetilde{Acc}_{g}\)}} 
& \multicolumn{1}{c|}{\cellcolor{white}\(Acc_{g}\)} & \multicolumn{1}{c|}{\cellcolor{white}\(Acc_{e}(\Delta_{mal})\)}
& \multicolumn{1}{c|}{\cellcolor{white}\(Acc_{g}\)} & \multicolumn{1}{c|}{\cellcolor{white}\(Acc_{e}(\Delta_{mal})\)}
& \multicolumn{1}{c|}{\cellcolor{white}\(Acc_{g}\)} & \multicolumn{1}{c|}{\cellcolor{white}\(Acc_{e}(\Delta_{mal})\)}
& \multicolumn{1}{c|}{\cellcolor{white}\(Acc_{g}\)} & \multicolumn{1}{c|}{\cellcolor{white}\(Acc_{e}(\Delta_{mal})\)}
& \multicolumn{1}{c|}{\cellcolor{white}\(Acc_{g}\)} & \multicolumn{1}{c|}{\cellcolor{white}\(Acc_{e}(\Delta_{mal})\)}
& \multicolumn{1}{c|}{\cellcolor{white}\(Acc_{g}\)} & \multicolumn{1}{c}{\cellcolor{white}\(Acc_{e}(\Delta_{mal})\)}\\
\hline\hline

& FedAvg\cite{mcmahan2017communication} & 98.60 &
98.17 & \multicolumn{1}{c|}{98.42(\textcolor[HTML]{33b872}{\textbf{+0.25}})} & 
97.73 & \multicolumn{1}{c|}{98.12(\textcolor[HTML]{33b872}{\textbf{+0.39}})} & 
98.43 & \multicolumn{1}{c|}{98.57(\textcolor[HTML]{33b872}{\textbf{+0.14}})} & 
97.96 & \multicolumn{1}{c|}{98.42(\textcolor[HTML]{33b872}{\textbf{+0.46}})} & 
98.48 & \multicolumn{1}{c|}{98.50(\textcolor[HTML]{33b872}{\textbf{+0.02}})}  & 
98.33 & \multicolumn{1}{c}{98.44(\textcolor[HTML]{33b872}{\textbf{+0.11}})}
\\
\rowcolor[HTML]{f2f2f2}
\cellcolor{white}&  FedProx\cite{li2020federated} & 98.62 &
98.20 & \multicolumn{1}{c|}{98.44(\textcolor[HTML]{33b872}{\textbf{+0.24}})} & 
97.76 & \multicolumn{1}{c|}{98.16(\textcolor[HTML]{33b872}{\textbf{+0.40}})} & 
98.22 & \multicolumn{1}{c|}{98.25(\textcolor[HTML]{33b872}{\textbf{+0.03}})} & 
98.07 & \multicolumn{1}{c|}{98.18(\textcolor[HTML]{33b872}{\textbf{+0.11}})} & 
98.51 & \multicolumn{1}{c|}{98.51(\textcolor[HTML]{33b872}{\textbf{+0}})}  & 
98.34 & \multicolumn{1}{c}{98.44(\textcolor[HTML]{33b872}{\textbf{+0.10}})}
\\
& Bulyan\cite{guerraoui2018hidden} & 98.64 &
98.01 & \multicolumn{1}{c|}{97.97(\textcolor[HTML]{e8967a }{\textbf{-0.04}})} & 
91.97 & \multicolumn{1}{c|}{97.98(\textcolor[HTML]{33b872}{\textbf{+6.01}})}&
98.46 & \multicolumn{1}{c|}{98.13(\textcolor[HTML]{e8967a }{\textbf{-0.33}})} & 
93.94 & \multicolumn{1}{c|}{98.08(\textcolor[HTML]{33b872}{\textbf{+4.14    }})} & 
98.01 & \multicolumn{1}{c|}{97.93(\textcolor[HTML]{e8967a}{\textbf{-0.08}})}  & 
97.51 & \multicolumn{1}{c}{97.61(\textcolor[HTML]{33b872}{\textbf{+0.10}})}
\\
\rowcolor[HTML]{f2f2f2}
\cellcolor{white}&  Multi-Krum\cite{blanchard2017machine} & 98.60 &
98.13 & \multicolumn{1}{c|}{98.30(\textcolor[HTML]{33b872}{\textbf{+0.17}})} & 
96.11 & \multicolumn{1}{c|}{98.06(\textcolor[HTML]{33b872}{\textbf{+1.95}})} & 
98.42 & \multicolumn{1}{c|}{98.53(\textcolor[HTML]{33b872}{\textbf{+0.11}})} & 
96.44 & \multicolumn{1}{c|}{98.08(\textcolor[HTML]{33b872}{\textbf{+1.64}})} & 
98.46 & \multicolumn{1}{c|}{98.28(\textcolor[HTML]{e8967a}{\textbf{-0.18}})}  & 
98.08 & \multicolumn{1}{c}{98.09(\textcolor[HTML]{33b872}{\textbf{+0.01}})}
\\
& Trimmed-mean\cite{pmlr-v80-yin18a} & 98.58 &
98.14 & \multicolumn{1}{c|}{98.20(\textcolor[HTML]{33b872}{\textbf{+0.06}})} & 
97.05 & \multicolumn{1}{c|}{98.06(\textcolor[HTML]{33b872}{\textbf{+1.01}})} & 
98.52 & \multicolumn{1}{c|}{98.13(\textcolor[HTML]{e8967a}{\textbf{-0.39}})} & 
97.65 & \multicolumn{1}{c|}{98.05(\textcolor[HTML]{33b872}{\textbf{+0.40}})} & 
98.54 & \multicolumn{1}{c|}{98.45(\textcolor[HTML]{e8967a}{\textbf{-0.09}})} & 
98.26 & \multicolumn{1}{c}{98.32(\textcolor[HTML]{33b872}{\textbf{+0.06}})}
\\
\rowcolor[HTML]{f2f2f2}
\multirow{-6}{*}{\cellcolor{white}\makecell{\textbf{MNIST} \\ \textbf{(CNN)}} }
&  Median\cite{pmlr-v80-yin18a} & 98.56 &
98.16 & \multicolumn{1}{c|}{98.44(\textcolor[HTML]{33b872}{\textbf{+0.28}})} &
96.21 & \multicolumn{1}{c|}{98.07(\textcolor[HTML]{33b872}{\textbf{+1.86}})} & 
98.16 & \multicolumn{1}{c|}{98.36(\textcolor[HTML]{33b872}{\textbf{+0.2}})} & 
97.78 & \multicolumn{1}{c|}{98.02(\textcolor[HTML]{33b872}{\textbf{+0.24}})} & 
97.75 & \multicolumn{1}{c|}{97.80(\textcolor[HTML]{33b872}{\textbf{+0.05}})} & 
97.61 & \multicolumn{1}{c}{97.75(\textcolor[HTML]{33b872}{\textbf{+0.14}})} 

\\
\hline

& FedAvg\cite{mcmahan2017communication} &  85.67 &
84.87 & \multicolumn{1}{c|}{85.11(\textcolor[HTML]{33b872}{\textbf{+0.24}})} & 
84.14 & \multicolumn{1}{c|}{84.87(\textcolor[HTML]{33b872}{\textbf{+0.73}})} & 
84.77 & \multicolumn{1}{c|}{84.92(\textcolor[HTML]{33b872}{\textbf{+0.15}})} & 
84.01 & \multicolumn{1}{c|}{84.92(\textcolor[HTML]{33b872}{\textbf{+0.91}})} &  
85.25 & \multicolumn{1}{c|}{85.25(\textcolor[HTML]{33b872}{\textbf{+0}})} & 
85.11 & \multicolumn{1}{c}{85.18(\textcolor[HTML]{33b872}{\textbf{+0.07}})} 
\\
\rowcolor[HTML]{f2f2f2} 
\cellcolor{white}& FedProx\cite{li2020federated} & 85.62 &
84.86 & \multicolumn{1}{c|}{85.04(\textcolor[HTML]{33b872}{\textbf{+0.18}})} & 
84.11 & \multicolumn{1}{c|}{84.93(\textcolor[HTML]{33b872}{\textbf{+0.82}})} & 
84.84 & \multicolumn{1}{c|}{84.96(\textcolor[HTML]{33b872}{\textbf{+0.12}})} & 
84.76 & \multicolumn{1}{c|}{84.92(\textcolor[HTML]{33b872}{\textbf{+0.16}})} & 
85.26 & \multicolumn{1}{c|}{85.30(\textcolor[HTML]{33b872}{\textbf{+0.04}})} & 
85.06 & \multicolumn{1}{c}{85.15(\textcolor[HTML]{33b872}{\textbf{+0.09}})} 
\\
& Bulyan\cite{guerraoui2018hidden} & 85.56 &
84.57 & \multicolumn{1}{c|}{84.61(\textcolor[HTML]{33b872}{\textbf{+0.04}})} & 
78.36 & \multicolumn{1}{c|}{84.08(\textcolor[HTML]{33b872}{\textbf{+5.72}})} & 
84.38 & \multicolumn{1}{c|}{84.60(\textcolor[HTML]{33b872}{\textbf{+0.22}})} & 
84.07 & \multicolumn{1}{c|}{84.45(\textcolor[HTML]{33b872}{\textbf{+0.38}})} & 
84.45 & \multicolumn{1}{c|}{84.30(\textcolor[HTML]{e8967a}{\textbf{-0.15}})} & 
84.09 & \multicolumn{1}{c}{84.22(\textcolor[HTML]{33b872}{\textbf{+0.13}})} 
\\
\rowcolor[HTML]{f2f2f2}
\cellcolor{white}&  Multi-Krum\cite{blanchard2017machine} & 85.68 &
84.82 & \multicolumn{1}{c|}{85.25(\textcolor[HTML]{33b872}{\textbf{+0.43}})} & 
81.05 & \multicolumn{1}{c|}{84.21(\textcolor[HTML]{33b872}{\textbf{+3.16}})} & 
84.69 & \multicolumn{1}{c|}{85.28(\textcolor[HTML]{33b872}{\textbf{+0.59}})} & 
82.26 & \multicolumn{1}{c|}{84.47(\textcolor[HTML]{33b872}{\textbf{+2.21}})} & 
84.64 & \multicolumn{1}{c|}{84.34(\textcolor[HTML]{e8967a}{\textbf{-0.30}})} & 
84.21 & \multicolumn{1}{c}{84.48(\textcolor[HTML]{33b872}{\textbf{+0.27}})} 
\\
& Trimmed-mean\cite{pmlr-v80-yin18a} & 85.67 &
84.64 & \multicolumn{1}{c|}{85.32(\textcolor[HTML]{33b872}{\textbf{+0.68}})} & 
83.09 & \multicolumn{1}{c|}{84.5(\textcolor[HTML]{33b872}{\textbf{+1.41}})} & 
84.54 & \multicolumn{1}{c|}{85.30(\textcolor[HTML]{33b872}{\textbf{+0.76}})} & 
83.73 & \multicolumn{1}{c|}{84.77(\textcolor[HTML]{33b872}{\textbf{+1.04}})} & 
84.55 & \multicolumn{1}{c|}{84.60(\textcolor[HTML]{33b872}{\textbf{+0.05}})} & 
84.41 & \multicolumn{1}{c}{84.48(\textcolor[HTML]{33b872}{\textbf{+0.07}})} 
\\
\rowcolor[HTML]{f2f2f2}
\cellcolor{white}\multirow{-6}{*}{\makecell{\textbf{FASHION} \\ \textbf{(CNN)}} }&  Median\cite{pmlr-v80-yin18a} & 85.06 &
84.75 & \multicolumn{1}{c|}{85.32(\textcolor[HTML]{33b872}{\textbf{+0.57}})} & 
82.68 & \multicolumn{1}{c|}{84.57(\textcolor[HTML]{33b872}{\textbf{+1.89}})} & 
84.16 & \multicolumn{1}{c|}{84.36(\textcolor[HTML]{33b872}{\textbf{+0.20}})} & 
84.71 & \multicolumn{1}{c|}{85.21(\textcolor[HTML]{33b872}{\textbf{+0.50}})} & 
84.22 & \multicolumn{1}{c|}{84.46(\textcolor[HTML]{33b872}{\textbf{+0.24}})} & 
83.41 & \multicolumn{1}{c}{84.08(\textcolor[HTML]{33b872}{\textbf{+0.67}})} 
\\
\hline

& FedAvg\cite{mcmahan2017communication} & 61.28 &
50.79 & \multicolumn{1}{c|}{53.05(\textcolor[HTML]{33b872}{\textbf{+2.26}})} & 
41.43 & \multicolumn{1}{c|}{52.24(\textcolor[HTML]{33b872}{\textbf{+10.81}})} & 
54.90 & \multicolumn{1}{c|}{57.55(\textcolor[HTML]{33b872}{\textbf{+2.65}})} & 
38.84 & \multicolumn{1}{c|}{55.99(\textcolor[HTML]{33b872}{\textbf{+17.15}})} &  
55.19& \multicolumn{1}{c|}{56.05(\textcolor[HTML]{33b872}{\textbf{+0.86}})} & 
54.40 & \multicolumn{1}{c}{56.47(\textcolor[HTML]{33b872}{\textbf{+2.07}})} 
\\
\rowcolor[HTML]{f2f2f2}
\cellcolor{white}&  FedProx\cite{li2020federated} & 61.50 &
51.21 & \multicolumn{1}{c|}{53.21(\textcolor[HTML]{33b872}{\textbf{+2.00}})} & 
42.04 & \multicolumn{1}{c|}{52.16(\textcolor[HTML]{33b872}{\textbf{+10.12}})} & 
52.16 & \multicolumn{1}{c|}{54.03(\textcolor[HTML]{33b872}{\textbf{+1.87}})} & 
35.40 & \multicolumn{1}{c|}{56.96(\textcolor[HTML]{33b872}{\textbf{+21.56}})} &  
56.00 & \multicolumn{1}{c|}{56.83(\textcolor[HTML]{33b872}{\textbf{+0.83}})} & 
55.01 & \multicolumn{1}{c}{57.1(\textcolor[HTML]{33b872}{\textbf{+2.09}})} 
\\
& Bulyan\cite{guerraoui2018hidden} & 59.42 &
46.27 & \multicolumn{1}{c|}{50.53(\textcolor[HTML]{33b872}{\textbf{+4.26}})} & 
16.23 & \multicolumn{1}{c|}{51.61(\textcolor[HTML]{33b872}{\textbf{+35.38}})} & 
52.96 & \multicolumn{1}{c|}{53.10(\textcolor[HTML]{33b872}{\textbf{+0.14}})} & 
22.18 & \multicolumn{1}{c|}{53.65(\textcolor[HTML]{33b872}{\textbf{+31.47}})} & 
54.06 & \multicolumn{1}{c|}{54.03(\textcolor[HTML]{e8967a}{\textbf{-0.03}})} & 
49.52 & \multicolumn{1}{c}{53.37(\textcolor[HTML]{33b872}{\textbf{+3.85}})} 
\\
\rowcolor[HTML]{f2f2f2}
\cellcolor{white}&  Multi-Krum\cite{blanchard2017machine} & 56.17 &
49.56 & \multicolumn{1}{c|}{52.52(\textcolor[HTML]{33b872}{\textbf{+2.96}})} & 
30.06 & \multicolumn{1}{c|}{51.78(\textcolor[HTML]{33b872}{\textbf{+21.72}})} & 
50.02 & \multicolumn{1}{c|}{50.18(\textcolor[HTML]{33b872}{\textbf{+0.16}})} & 
34.58 & \multicolumn{1}{c|}{51.84(\textcolor[HTML]{33b872}{\textbf{+17.26}})} & 
53.09 & \multicolumn{1}{c|}{53.61(\textcolor[HTML]{33b872}{\textbf{+0.52}})} & 
51.05 & \multicolumn{1}{c}{53.01(\textcolor[HTML]{33b872}{\textbf{+1.96}})} 
\\
& Trimmed-mean\cite{pmlr-v80-yin18a} & 56.27 &
46.67 & \multicolumn{1}{c|}{49.70(\textcolor[HTML]{33b872}{\textbf{+3.03}})} & 
35.66 & \multicolumn{1}{c|}{51.88(\textcolor[HTML]{33b872}{\textbf{+16.22}})} & 
48.21 & \multicolumn{1}{c|}{52.25(\textcolor[HTML]{33b872}{\textbf{+4.04}})} & 
30.09 & \multicolumn{1}{c|}{50.93(\textcolor[HTML]{33b872}{\textbf{+20.84}})} & 
52.88 & \multicolumn{1}{c|}{53.27(\textcolor[HTML]{33b872}{\textbf{+0.39}})} & 
51.55 & \multicolumn{1}{c}{52.80(\textcolor[HTML]{33b872}{\textbf{+1.25}})} 
\\
\rowcolor[HTML]{f2f2f2} 
\cellcolor{white}
\multirow{-6}{*}{\makecell{\textbf{CIFAR-10} \\ \textbf{(Alexnet)}} }
&  Median\cite{pmlr-v80-yin18a} & 55.25 &
49.43 & \multicolumn{1}{c|}{50.90(\textcolor[HTML]{33b872}{\textbf{+1.47}})} & 
31.64 & \multicolumn{1}{c|}{51.69(\textcolor[HTML]{33b872}{\textbf{+20.05}})} & 
47.94 & \multicolumn{1}{c|}{51.42(\textcolor[HTML]{33b872}{\textbf{+3.48}})} & 
29.19 & \multicolumn{1}{c|}{49.02(\textcolor[HTML]{33b872}{\textbf{+19.83}})} &  
51.08 & \multicolumn{1}{c|}{52.03(\textcolor[HTML]{33b872}{\textbf{+0.95}})} & 
43.94 & \multicolumn{1}{c}{51.36(\textcolor[HTML]{33b872}{\textbf{+7.42}})} \\
\hline

\end{tabular} 
}
\end{table*}

\subsection{Comparison Experiments}

\subsubsection{Global vs. Adversarial Local Performance}

To measure the impact and utility of FedThief, we compare the accuracy of the global model with that of the malicious clients’ ensemble model across a variety of attack-defense scenarios, as summarized in Table~\ref{tab:acc}.

Across practically all configurations, results reveal that the malicious clients consistently gain access to models with better accuracy than the shared global model. This advantage confirms the success of the SCFL paradigm introduced by FedThief. Importantly, this benefit is achieved \textit{without} reducing the poisoning strength of the underlying attack. That is, the global model still suffers substantial performance degradation due to malicious behavior, yet malicious clients themselves retain high model utility.

Further, the severity of global model degradation strongly correlates with the proportion of adversarial clients. For instance, when \(\alpha = 0.4\), malicious clients are capable of reducing global accuracy by more than 20\%, 
while maintaining their personal performance levels through ensemble optimization. This demonstrates the asymmetric nature of adversarial benefits facilitated by FedThief.

\begin{figure*}[ht]
    \centering
    \setlength{\abovecaptionskip}{-5pt}
    \setlength{\belowcaptionskip}{0pt}
    \includegraphics[width=0.9\textwidth]{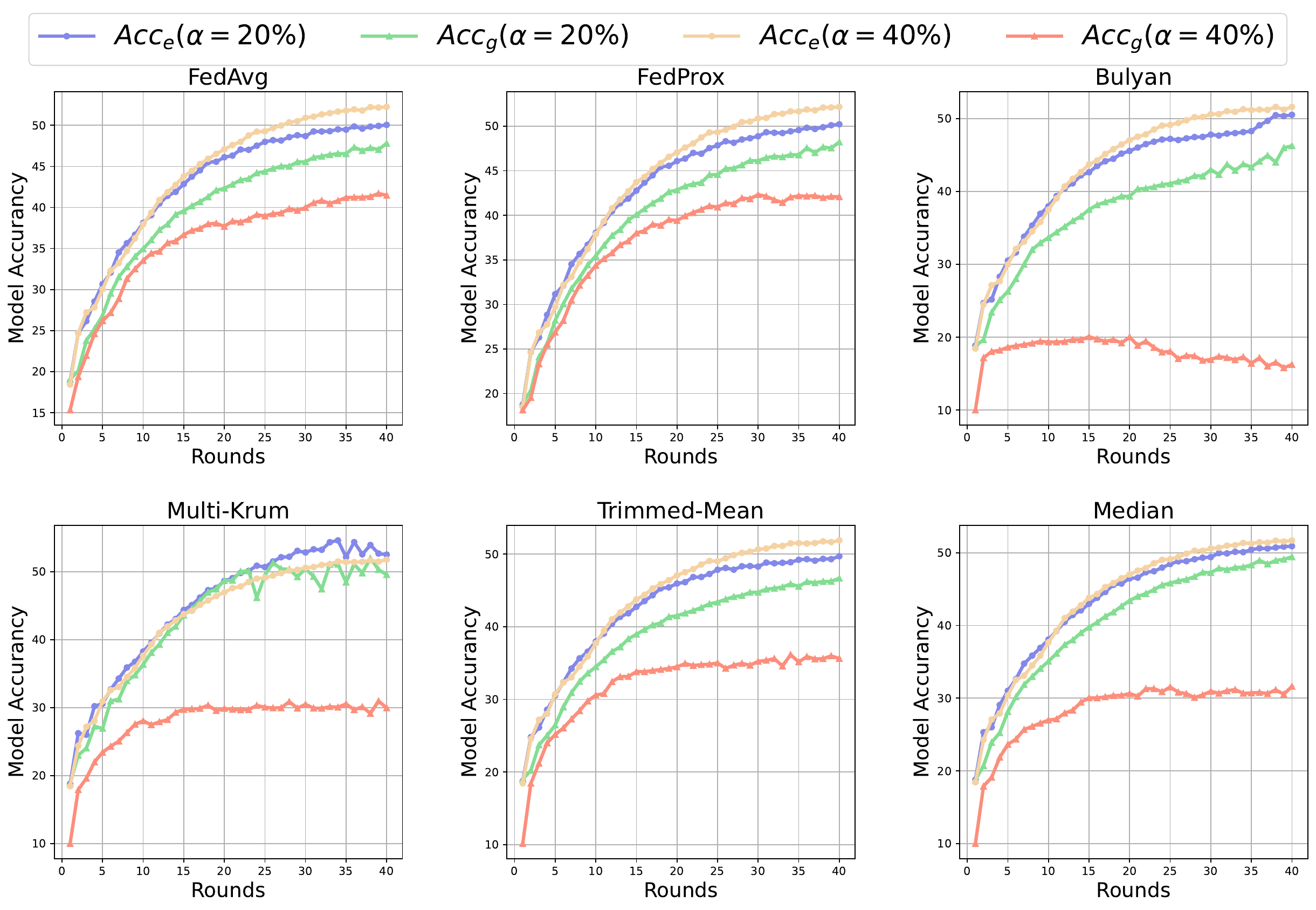}
    \caption{Accuracy evolution of malicious ensembles vs. global model under the LIE attack on CIFAR-10, with malicious ratios \(\alpha = 0.2\) and \(\alpha = 0.4\). Six defense aggregation schemes are included.}
    \label{fig:acc}
\end{figure*}

To better visualize the performance gap over time, we plot the training progress under the LIE attack on CIFAR-10 in Fig.~\ref{fig:acc}. The figure clearly illustrates that malicious clients achieve higher accuracy from the early training stages, and this advantage persists throughout all communication rounds and across different defensive aggregation strategies.

Interestingly, this adversarial advantage is more pronounced in complex datasets (e.g., CIFAR-10 or FASHION-MNIST), while relatively tame in simpler ones like MNIST. This is likely due to the lower representation capacity needed to converge on MNIST, making its models less sensitive to gradient manipulations and knowledge expansion via distillation. In contrast, more complex tasks benefit substantially from ensemble-guided optimization, providing malicious clients with a tangible private-model boost.

\textbf{Generalization to Data Poisoning Attacks.}  
While most experimental evaluations focus on model poisoning through manipulated gradients, we further test the generality of FedThief under label-level data poisoning scenarios. Specifically, we conduct attacks using two representative label-flipping strategies: \emph{Symmetry Flip}, where all class labels are cyclically permuted; and \emph{Pairwise Flip}, where each label is flipped to its adjacent class. These attacks differ fundamentally from gradient-based Byzantine behavior, as they alter training labels rather than update directions.

As demonstrated in Table~\ref{tab:I}, FedThief remains effective across both attack types. In all datasets tested (MNIST, FASHION, CIFAR-10), the ensemble models trained by malicious clients outperform the global model, even though the training process introduces significant label noise. The advantage is most prominent on CIFAR-10, where data complexity provides greater benefit to ensemble-guided optimization. For example, under the Pair Flip attack on MNIST, malicious clients achieve 98.06\% accuracy with FedThief, compared to only 51.50\% by the corrupted global model.

\begin{table}[t]
\caption{
Performance of FedThief under data poisoning attacks with Bulyan aggregation ($\alpha=0.4$). Each entry shows global and ensemble accuracy, respectively.
}
\label{tab:I}
\centering
{
\begin{tabular}{c||c|c|c|c}
\hline \thickhline

\rowcolor{mygray}
  & \multicolumn{2}{c|}{Symmetry Flipping} 
  & \multicolumn{2}{c}{Pair Flipping} \\

\cline{2-5}
\rowcolor{mygray}
  \multirow{-2}{*}{\cellcolor{mygray}\centering Dataset} 
  & \multicolumn{1}{c|}{\(Acc_{g}\)} & \multicolumn{1}{c|}{\(Acc_{e}(\Delta_{mal})\)}
  & \multicolumn{1}{c|}{\(Acc_{g}\)} & \multicolumn{1}{c}{\(Acc_{e}(\Delta_{mal})\)}\\
\hline
MNIST
  & 97.28 & \multicolumn{1}{c|}{98.05(\textcolor[HTML]{33b872}{\textbf{+0.77}})} & 
51.50 & \multicolumn{1}{c}{98.06(\textcolor[HTML]{33b872}{\textbf{+46.56}})} \\
\hline
FASHION
&
81.44 & \multicolumn{1}{c|}{84.43(\textcolor[HTML]{33b872}{\textbf{+2.99}})} 
& 41.84 & \multicolumn{1}{c}{83.69(\textcolor[HTML]{33b872}{\textbf{+41.85}})} 
\\ 
\hline
CIFAR10     
  &
  39.75 & \multicolumn{1}{c|}{49.69(\textcolor[HTML]{33b872}{\textbf{+9.94}})}
  & 26.12 & \multicolumn{1}{c}{49.42(\textcolor[HTML]{33b872}{\textbf{+23.30}})} 
 \\ 
\hline
\end{tabular}}
\end{table}

This experimental evidence suggests that FedThief is agnostic to the specific form of the attack vector—whether in parameter space or label domain—and highly capable of extending to a broader class of adversarial strategies beyond model poisoning. The inherent resilience and dynamic adaptability of the ensemble approach underlie its broader applicability in practical federated environments where attack methodologies may evolve or diversify.

\subsubsection{Client Utility Spectrum}

To rigorously characterize the differential impacts of FedThief, we propose a framework that establishes two fundamental performance limits. The theoretical lower bound is defined by \( Acc_{\text{local}} \), which captures the baseline model accuracy achievable through completely isolated local training without any federated knowledge transfer. The theoretical upper bound is given by \( \widetilde{Acc}_g \), representing the maximum attainable accuracy of a global model that remains uncompromised by any adversarial attacks, as FedThief's knowledge acquisition does not exceed what would be obtained through normal federation.

\begin{figure}[ht]
    \centering
    \includegraphics[width=0.48\textwidth]{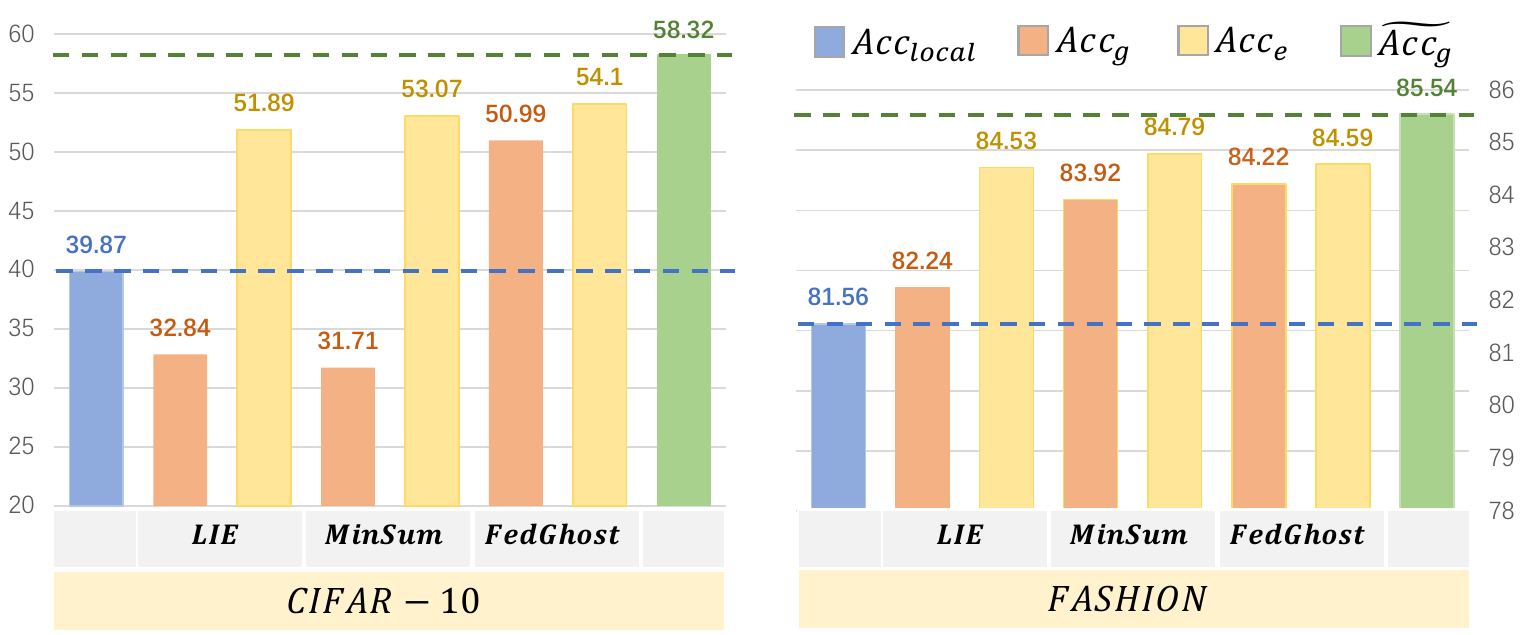}
    \caption{
Expected accuracy for malicious and benign clients under the FedThief framework, evaluated on CIFAR-10 and Fashion-MNIST datasets with malicious client ratio $\alpha = 0.4$.
    }
    \label{fig:benifit}
    \vspace{-3pt}
\end{figure}

Given that clients are unaware of the server's aggregation rule or the global attack state, we compute the \textit{expected} accuracy each client type (benign or malicious) would perceive as the average across all evaluated aggregation schemes. As illustrated in Fig.~\ref{fig:benifit}, this analysis shows that malicious clients consistently exceed both \( Acc_{\text{local}} \) and collaborative expectations, highlighting the personal gain facilitated by FedThief. Conversely, benign clients often fall below their local baselines, indicating that their collaboration in a poisoned federation may actively harm their personal training outcome.

These findings expose a critical fairness breach in federated learning: malicious clients can strategically exploit collaborativity, significantly outperforming honest users, and undermining the equality of benefits in the system.

\subsection{Ablation Study}

We perform an extensive ablation study to rigorously analyze the individual contributions of FedThief's core components and systematically evaluate its critical design parameters.

\subsubsection{Preservation of Attack Behavior}

A central design principle in FedThief is that it should not interfere with attack behavior. To confirm this property, we compare the global accuracy \( Acc_{g} \) under traditional Byzantine attacks (e.g., LIE) with and without FedThief. Since only the malicious model \(\theta_m\) is used when generating gradient updates, FedThief does not alter the attack logic. As confirmed in Table~\ref{tab:acc}, attack degradation on the global model is identical, validating that FedThief preserves the underlying attack strength while adding private-model optimization.

\subsubsection{Effectiveness of Ensemble Optimization}

Ensemble optimization in FedThief is driven by a weighted loss function incorporating two terms: (i) cross-entropy on the private dataset, and (ii) KL divergence from ensemble-generated soft targets. The trade-off between these is controlled via a weight \(\lambda\). 
We investigate the impact of varying the regularization parameter $\lambda \in \{0, 0.25, 0.5, 0.75, 1.0\}$ on model performance through extensive experiments conducted on CIFAR-10.

\begin{figure}[t]
    \centering
    \setlength{\abovecaptionskip}{-5pt}
    \setlength{\belowcaptionskip}{5pt}
    \includegraphics[width=0.48\textwidth]{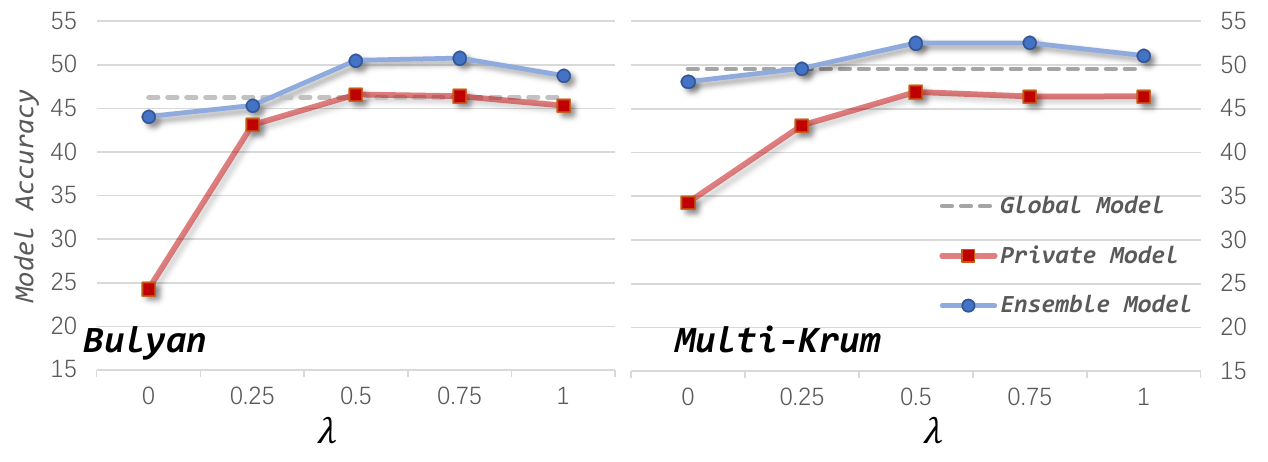}
    \caption{Performance under different settings of loss balance factor \(\lambda\), evaluated on CIFAR-10 with \(\alpha = 0.2\).}
    \label{fig:lamda}
\end{figure}

As plotted in Fig.~\ref{fig:lamda}, using only cross-entropy loss (\(\lambda = 1\)) results in limited generalization, especially for clients with small or biased data. Lower values of \(\lambda\) allow the ensemble to guide training, improving both local and ensemble model accuracy. However, too small a \(\lambda\) distorts the optimization process—leading to convergence toward the global model, losing private uniqueness. The optimal model performance is observed at $\lambda = 0.5$, demonstrating that an intermediate trade-off between the competing objectives yields favorable results.

\subsubsection{Impact of Validation Partition Ratio}

Parameter \( v \) controls how the private dataset is split between training the malicious model and validating the ensemble. A lower \( v \) allocates more data to validation (benefiting the ensemble), while larger \( v \) favors attack model training.

Table~\ref{tab:v} presents performance results under various \(v\) values. A moderate split ratio of \(v=5\) yields the best balance, enabling both high-quality ensemble optimization and sufficient attack gradient calculation. Too aggressive splitting (\( v=10 \)) weakens the ensemble’s generalization capacity, whereas too little validation (\( v=2 \)) hampers adversarial manipulation strength.

\begin{table}[t]\small
\caption{Impact of the private dataset partition ratio \( v \), evaluated on CIFAR-10 with \(\alpha = 0.2\). Attack: LIE. Aggregators: Bulyan and Multi-Krum.}
\label{tab:v}
\centering
\scriptsize 
\resizebox{\linewidth}{!}{ 
\setlength\tabcolsep{3pt}
\renewcommand\arraystretch{1.2}
\begin{tabular}{r|c|c|c|c|c|c}
\hline\thickhline
\rowcolor[HTML]{f2f2f2}
& \multicolumn{3}{c|}{FedBulyan} 
& \multicolumn{3}{c}{Multi-Krum} \\
\rowcolor[HTML]{f2f2f2}
& $v=2$ & $v=5$ & $v=10$ 
& $v=2$ & $v=5$ & $v=10$ \\
\hline
\rowcolor[HTML]{f9f9f9} \(Acc_{g}\) & 49.09 & 46.27 & 47.01 & 51.57 & 49.56 & 49.83 \\  
\rowcolor[HTML]{ffffff} \(Acc_{e}\) & 50.92 & 50.53 & 50.79 & 53.39 & 52.52 & 51.79 \\  
\rowcolor[HTML]{f9f9f9} \(\Delta_{\text{mal}}\) & \textcolor[HTML]{33b872}{+1.83}  & \textcolor[HTML]{33b872}{+4.26} & \textcolor[HTML]{33b872}{+3.78} & \textcolor[HTML]{33b872}{+1.82} & \textcolor[HTML]{33b872}{+2.96} & \textcolor[HTML]{33b872}{+1.96} \\  
\hline
\end{tabular} 
}
\end{table}

\subsubsection{Effectiveness of Ensemble Components}

To better understand the internal mechanism of FedThief’s ensemble-guided optimization, we conduct a detailed component-wise analysis on the ensemble structure. Specifically, we investigate the individual and joint contributions of three distinct model sources—namely, the private model $\theta_p$ trained on the adversarial client’s local dataset, the malicious model $\theta_m$ synchronized with the global model for gradient poisoning, and the error model $\theta_e$ which approximates the residual discrepancy between $\theta_m$ and the global aggregation output.

Each of these models encodes different inductive biases and information flows during the adversarial training process. $\theta_p$ captures client-specific patterns, biases, and unique local semantic distributions. $\theta_m$ reflects the poisoned global trajectory, implicitly embedding the impact of attack-induced updates across communication rounds. $\theta_e$ serves as a corrective signal that captures counterfactual behavior—representing the divergence introduced through adversarial manipulation.

Table~\ref{tab:ensemble_ablation} reports the ensemble model accuracy under various inclusion settings across three benchmark datasets: MNIST, FASHION, and CIFAR-10. When only $\theta_p$ is used, the ensemble corresponds to the naive baseline relying solely on private data. Adding $\theta_m$ to the ensemble consistently improves accuracy, suggesting that tracking global model dynamics—even in poisoned form—provides additional and useful guidance during model distillation. Interestingly, including $\theta_e$ alone with $\theta_p$ yields similar improvements, particularly on more complex datasets such as FASHION or CIFAR-10. This empirical result implies that modeling error residuals introduces high-frequency correction signals that can enhance generalization beyond client-local or global knowledge alone.

The full ensemble, incorporating all three model sources, achieves the highest accuracy across all datasets. The most notable gains are observed on CIFAR-10, where the inclusion of $\theta_m$ and $\theta_e$ contributes 2.5\% improvement over the private-only ensemble. These results indicate that the full ensemble benefits from a multi-view knowledge composition, combining local specificity, global drift, and adversarial perturbation effects into a unified predictive output. Such diversity appears essential for optimal private optimization in the presence of adversarial gradients and federated heterogeneity.

This component-wise ablation confirms that model composition is critical in harnessing the full power of ensemble-based knowledge distillation under adversarial settings. It further strongly supports the core design intuition that combining orthogonal information sources synergistically enhances both robustness and transferability in SCFL.

\begin{table}[t]
\caption{Accuracy of the ensemble model when including each component. \textcolor[HTML]{33b872}\checkmark means the component is included. Min-Sum is used under FedAvg aggregation.}
\centering
{
\begin{tabular}
{
% @{\hspace{1pt}}
c
@{\hspace{4pt}}
c
@{\hspace{4pt}}
c
|c|c|c|c|c|c}
\hline \thickhline
\rowcolor{mygray}
&&&
\multicolumn{2}{c|}{MNIST}
&
\multicolumn{2}{c|}{FASHION}
&
\multicolumn{2}{c}{CIFAR10}\\

\rowcolor{mygray}
\multirow{-2}{*}{$\theta_p$} 
&
\multirow{-2}{*}{$\theta_m$}  
&
\multirow{-2}{*}{$\theta_e$}  
&
$\alpha${\tiny$=$}$0.2$ & $\alpha${\tiny$=$}$0.4$ & $\alpha${\tiny$=$}$0.2$ & $\alpha${\tiny$=$}$0.4$ & $\alpha${\tiny$=$}$0.2$ & $\alpha${\tiny$=$}$0.4$ \\
\hline
\textcolor[HTML]{33b872}\checkmark 
& & & 98.30 & 98.03 & 82.85 & 83.11 & 49.27 & 50.18 
\\ 
\rowcolor[HTML]{f2f2f2}\textcolor[HTML]{33b872}\checkmark
&
\textcolor[HTML]{33b872}\checkmark  
& & 98.43 & 98.06 & 83.61 & 83.17 & 54.62 & 52.99 \\ 
\textcolor[HTML]{33b872}\checkmark
& 
& 
\textcolor[HTML]{33b872}\checkmark 
& 98.16 & 97.85 & 81.49 & 82.01 & 52.28 & 53.06  \\ 
\rowcolor[HTML]{f2f2f2}\textcolor[HTML]{33b872}\checkmark
&
\textcolor[HTML]{33b872}\checkmark
& 
\textcolor[HTML]{33b872}\checkmark & 
98.57 & 98.42 & 84.92 &  84.92 &  57.55 & 55.99\\ 
\hline
\end{tabular}}
\label{tab:ensemble_ablation}
\end{table}

\subsubsection{Effect of Temperature in Distillation}

In knowledge distillation, the temperature hyperparameter \(T\) plays a critical role in shaping the softness of the teacher model's output distribution. Specifically, a higher temperature flattens the softmax probabilities, exposing the dark knowledge embedded in less confident class predictions, while a lower temperature sharpens the distribution to resemble one-hot targets. This spectrum of logit softening directly influences how effectively the student model---in this case, the privately distilled model on each malicious client---can align with the ensemble teacher.

To explore this effect within the FedThief framework, we conduct a distillation temperature ablation study on CIFAR-10 using the MinSum attack and FedAvg aggregation, fixing $\alpha=0.2$. The tested range includes \(T \in \{1, 2, 3, 5\}\), where \(T=1\) corresponds to standard softmax and higher values represent increasingly softer logits.

As presented in Table~\ref{tab:temp}, increasing the temperature from 1 to 3 improves ensemble-model accuracy by over 2\%, demonstrating the benefits of softened supervision for stabilizing optimization and enhancing generalization under non-identical logit sources. At \(T = 3\), the ensemble achieves its peak performance, validating this value as an empirically optimal point for knowledge transfer in our SCFL scenario. Notably, further increasing the temperature to \(T = 5\) does not result in continued improvement, indicating a saturation effect. When the output becomes overly uniform, the signal-to-noise ratio in the soft targets may degrade, weakening the gradient alignment between teacher and student models.

These findings align with prior distillation literature, which suggests that moderate temperature scaling helps reveal richer inter-class relationships while maintaining discriminability. In our context, it also supports better cross-model mimicry, allowing the malicious client to integrate private, poisoned, and global logics effectively into its local model. Thus, tuning the temperature hyperparameter is essential for unlocking the full potential of ensemble-guided optimization in adversarial federated learning.

\begin{table}[t]
\caption{Ensemble accuracy under different distillation temperatures on CIFAR-10 (MinSum, FedAvg, $\alpha=0.2$).}
\centering
\label{tab:temp}
\begin{tabular}{c|cccc}
\toprule
\textbf{Temperature} $T$ & 1 & 2 & 3 & 5 \\
\midrule
Accuracy (\%) & 55.12 & 57.30 & \textbf{57.55} & 57.53 \\
\bottomrule
\end{tabular}
\end{table}

%% file: sec/5_conclusion.tex
\section{Conclusion}
\label{sec:Conclusion}

This work presents a novel federated learning attack paradigm termed Self-Centered Federated Learning (SCFL), wherein malicious clients can simultaneously degrade the global model’s performance while improving their personal utility. We introduce FedThief, an instantiation of SCFL, which strategically separates global manipulation and local optimization via ensemble-guided training and adaptive model fusion. The framework leverages knowledge distillation and meta-predictive integration to jointly achieve global disruption and private benefit. Extensive experimental evaluations across diverse settings demonstrate the effectiveness, adaptability, and resilience of the proposed approach under varying levels of data heterogeneity and adversarial participation.

\textbf{Discussion and Future Directions.} While empirically effective, the proposed FedThief framework also presents several practical challenges. First, its ensemble-based training introduces non-trivial computational and communication overhead for malicious clients. Although this is feasible for resourceful adversaries, reducing the cost while preserving attack efficacy remains an open issue. Second, the effectiveness of SCFL may be influenced by the availability and quality of private data on malicious clients. Since ensemble alignment and local distillation rely on locally-held samples, data that is highly imbalanced, sparse, or non-representative—particularly under non-IID distributions—could reduce knowledge transfer efficiency and slightly diminish the private utility gained. Nonetheless, in practical scenarios, malicious clients may still possess sufficient data diversity to sustain meaningful benefit. To expand applicability, future work could explore methods such as data synthesis or adversarial augmentation to mitigate the impact of limited or biased data.

Future research may explore lightweight attack formulations with reduced model complexity and fewer communication rounds. Moreover, developing adaptive mechanisms—such as self-supervised objectives or on-device data augmentation—to mitigate data scarcity could further enhance robustness under realistic deployment scenarios. From a defensive viewpoint, this work highlights the need for aggregation strategies resilient to utility-driven adversaries. Advancing secure and adaptive aggregation schemes that can detect and counter such stealthy behaviors remains an important direction for building trustworthy federated systems.